\documentclass{ieeeaccess}

\usepackage{cite}
\usepackage{amsmath,amssymb,amsfonts}
\usepackage{textcomp}

\usepackage{graphicx}
\usepackage[ruled,vlined]{algorithm2e}
\usepackage{hyperref}
\usepackage{float}
\usepackage{multirow}
\usepackage{booktabs,caption}
\usepackage[flushleft]{threeparttable}
\usepackage{subcaption}
\floatstyle{plaintop}
\restylefloat{table}

\usepackage{fancyhdr}
\usepackage{kantlipsum}
\usepackage{eso-pic}

\def\BibTeX{{\rm B\kern-.05em{\sc i\kern-.025em b}\kern-.08em
    T\kern-.1667em\lower.7ex\hbox{E}\kern-.125emX}}

\begin{document}

\AddToShipoutPictureBG*{%
	\AtPageLowerLeft{%
		\setlength\unitlength{1in}%
		\hspace*{\dimexpr0.5\paperwidth\relax}%%  change \dimexpr0.5\paperwidth\relax appropriately
		\makebox(0,0.63)[c]{2169-3536~\copyright2024 IEEE. This work has been accepted for publication in IEEE ACCESS.}
		\makebox(0,0.3)[c]{The published version can be accessed at \href{https://doi.org/10.1109/ACCESS.2024.3385122}{https://doi.org/10.1109/ACCESS.2024.3385122}.}
}}

\history{Date of publication xxxx 00, 0000, date of current version xxxx 00, 0000.}
\doi{10.1109/ACCESS.2024.3385122}

\title{DeepIPC: Deeply Integrated Perception and Control for an Autonomous Vehicle in Real Environments}
\author{\uppercase{Oskar Natan}\authorrefmark{1}, \IEEEmembership{Member, IEEE} and
\uppercase{Jun Miura}\authorrefmark{2}
\IEEEmembership{Member, IEEE}}

\address[1]{Department of Computer Science and Electronics, Universitas Gadjah Mada, Yogyakarta 55281 Indonesia (e-mail: oskarnatan@ugm.ac.id)}
\address[2]{Department of Computer Science and Engineering, Toyohashi University of Technology, Aichi 441-8580 Japan (e-mail: jun.miura@tut.jp)}

%\tfootnote{This paragraph of the first footnote will contain support
%information, including sponsor and financial support acknowledgment. For
%example, ``This work was supported in part by the U.S. Department of
%Commerce under Grant BS123456.''}

\markboth
{{\bf O. Natan and J. Miura}: DeepIPC: Deeply Integrated Perception and Control for an Autonomous Vehicle in Real Environments}
{{\bf O. Natan and J. Miura}: DeepIPC: Deeply Integrated Perception and Control for an Autonomous Vehicle in Real Environments}

\corresp{Corresponding author: Oskar Natan (e-mail: oskarnatan@ugm.ac.id).}

\begin{abstract}
%We propose DeepIPC, an end-to-end autonomous driving model that handles both perception and control tasks in driving a vehicle. The model consists of two main parts, perception and controller modules. The perception module takes an RGBD image to perform semantic segmentation and bird's eye view (BEV) semantic mapping along with providing their encoded features. Meanwhile, the controller module processes these features with the measurement of GNSS locations and angular speed to estimate waypoints that come with latent features. Then, two different agents are used to translate waypoints and latent features into a set of navigational controls to drive the vehicle. The model is evaluated by predicting driving records and performing automated driving under various conditions in real environments. The experimental results show that DeepIPC achieves the best drivability and multi-task performance even with fewer parameters compared to the other models. Codes and data will be published at \href{https://github.com/oskarnatan/DeepIPC}{https://github.com/oskarnatan/DeepIPC}.

In this work, we introduce DeepIPC, a novel end-to-end model tailored for autonomous driving, which seamlessly integrates perception and control tasks. Unlike traditional models that handle these tasks separately, DeepIPC innovatively combines a perception module, which processes RGBD images for semantic segmentation and generates bird's eye view (BEV) mappings, with a controller module that utilizes these insights along with GNSS and angular speed measurements to accurately predict navigational waypoints. This integration allows DeepIPC to efficiently translate complex environmental data into actionable driving commands. Our comprehensive evaluation demonstrates DeepIPC's superior performance in terms of drivability and multi-task efficiency across diverse real-world scenarios, setting a new benchmark for end-to-end autonomous driving systems with a leaner model architecture. The experimental results underscore DeepIPC's potential to significantly enhance autonomous vehicular navigation, promising a step forward in the development of autonomous driving technologies. For further insights and replication, we will make our code and datasets available at \href{https://github.com/oskarnatan/DeepIPC}{https://github.com/oskarnatan/DeepIPC}.

\end{abstract}


\titlepgskip=-21pt

\maketitle

\section{Introduction} \label{sec:intro}

End-to-end learning has become a preferable approach in autonomous driving as manual configuration to integrate task-specific modules is no longer needed. This technique allows the model to share useful features directly from perception modules to controller modules. Moreover, the model can learn and receive extra supervision from a multi-task loss function that considers several performance criteria. All these benefits result in a better model performance even with a smaller model size due to its compactness \cite{oskar_tiv}\cite{e2e_av2}. To date, there have been a lot of works in the field of end-to-end autonomous driving, whether it is based on simulation \cite{sim_av1}, or offline real-world where the model predicts a set of driving records \cite{offline_av}, or online real-world where the model is deployed for automated driving \cite{real_av}. Besides dealing with diverse conditions such as driving on a sunny day or low-light evening, another challenge that remains in online real-world autonomous driving is that the model must deal with noise and inaccuracy of sensor measurement. This issue needs to be addressed as it affects model performance and its generalization capability \cite{challenge1}\cite{teti}.

\begin{figure}[t]
	\begin{center}
		\includegraphics[width=\linewidth]{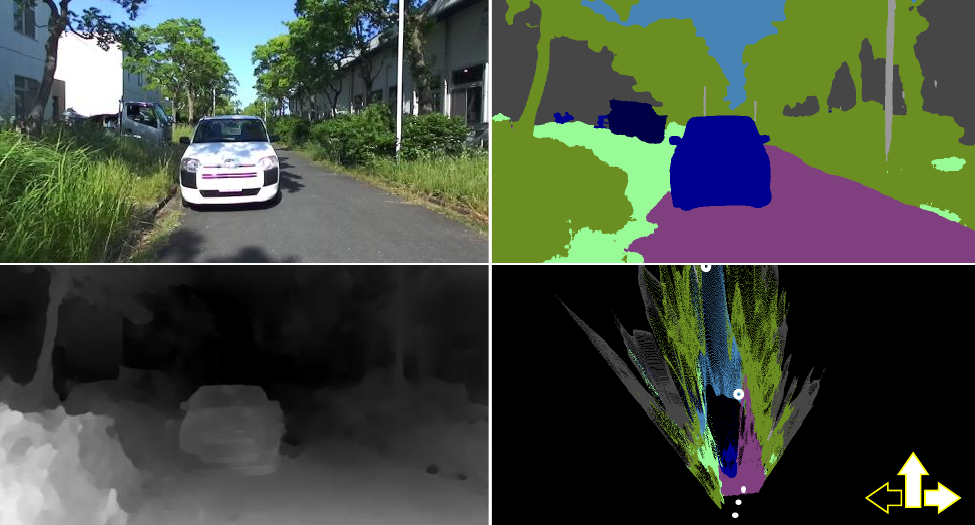}
		
	\end{center}
	%\vspace{-3mm}
	\caption{DeepIPC perceives the environment by performing image segmentation and BEV semantic mapping. Simultaneously, it also estimates waypoints (white dots) and controls to drive the vehicle by following a set of route points (white hollow circles). The detailed architecture of DeepIPC can be seen in Fig. \ref{fig:model}.}
	\label{fig:overview}
	%\vspace{-5mm}
\end{figure}

To address those challenges, some works have been conducted with a focus on simulation-to-real adaptation. Although the models still suffer from performance losses due to sensor inaccuracies and diverse conditions in real environments, these approaches are said to be promising for future autonomous driving \cite{sim2real3}\cite{sim2real4}. On the other hand, some different approaches have been proposed with a focus on end-to-end imitation learning where the model is trained to mimic an expert driver in dealing with complicated situations on the street \cite{imit2}\cite{imit3}. These approaches are preferred as they are easier and can be done with a simple supervised learning technique \cite{argall}. Moreover, plenty of publicly available datasets along with self-made datasets can be used to train the model to enrich its driving experiences. In a comprehensive review by Teng et al. \cite{teng_review}, various motion planning techniques for autonomous driving are evaluated, highlighting the progression toward more integrated perception-action systems. These methodologies underscore the evolving landscape of intelligent vehicle navigation, offering a nuanced understanding of how advanced planning strategies can inform perception-action coupling mechanisms. Conversely, Xu et al. \cite{xu_uncertain} detail a pioneering approach to uncertainty-aware exploration in robotics, presenting parallels in adaptive decision-making that are critical for autonomous vehicles. Their work elucidates the growing emphasis on robust perception-driven actions underpinned by sophisticated localization and mapping techniques. Considering its advantage, we adopt these approaches and propose a model namely DeepIPC (Deeply Integrated Perception and Control). The architecture of DeepIPC is based on our previous work \cite{oskar_tiv} with some improvements to deal with the issues. 

Concisely, DeepIPC is a model that can be forced to learn how to compensate for noise and inaccuracy of sensor measurement implicitly by mimicking expert behavior to achieve human-like autonomous driving \cite{human_like0}\cite{huifang}. DeepIPC processes multi-modal data that contain several quantities needed to perceive the environment and drive the vehicle in one forward pass. The perception parts take an RGBD image to perform semantic segmentation and BEV semantic mapping. Simultaneously, the controller parts estimate waypoints and navigational controls based on the extracted perception features, wheel's angular speeds, and route points. Unlike in an ideal simulated environment, DeepIPC must deal with real implementation issues. For example, it must compensate for the issue of inaccurate route points positioning caused by the inaccuracy of the GNSS receiver and IMU sensor. Then, there are also plenty of noises on the RGBD camera that affect the scene understanding capability. As shown in Fig. \ref{fig:overview}, DeepIPC must be able to safely avoid the obstacle by predicting navigational controls and waypoints correctly in the traversable area although the given route points (two white circles on the bottom-right image) are not located accurately in the local coordinate. Hence, we state the novelties as follows:
\begin{itemize}
	%\item We propose DeepIPC, an improved version of our previous model \cite{oskar_tiv} for driving a robotic vehicle in real environments. Concisely, the perception parts are modified to process a wider RGBD image for better perception. Then, the controller parts are enhanced with an updated control policy to allow a higher degree of maneuverability. Moreover, it is also fed with two route points and wheel's angular speeds for better driving intuition.
	\item We propose DeepIPC, which significantly advances our prior model \cite{oskar_tiv} by employing broader RGBD image analytics, enabling finer environmental perception. Enhanced control mechanisms offer augmented maneuverability, leveraging dual route points and nuanced wheel speed insights, facilitating more intuitive and responsive driving behaviors.
	
	%\item As we demonstrate the real-world experiments of our previous study on simulation-based autonomous driving \cite{oskar_tiv}, we exhibit how a proof-of-concept study from an ideal simulated environment can be realized by addressing some implementation issues. Furthermore, unlike many studies that only predict driving records, we deploy DeepIPC to perform real-world automated driving. This also exposes the usefulness of end-to-end imitation learning, not only for a simple model but also for a complex multi-input multi-output model. %Then, we also define a different evaluation metric where the best drivability is defined by the lowest driver intervention, not by the lowest number of collisions. This is necessary to avoid any damage to the vehicle during evaluation. %Unlike in an ideal simulated environment, we must deal with several issues such as sensor inaccuracy and noise.% 
	\item We demonstrate real-world deployment of DeepIPC, showing the practical translation from simulated proofs of concept to on-road applications, illustrating sophisticated end-to-end imitation learning's efficacy in navigating a multifaceted sensorimotor landscape, highlighting its versatility beyond mere driving record prediction.
	
	%\item We compare DeepIPC with other models to get a clearer performance justification. All models are evaluated by predicting a set of driving records and performing real-world automated driving under different conditions. Then, we also introduce a new evaluation metric to justify the performance of the models based on the number of driver interventions. This is necessary to avoid any collisions that can damage the vehicle. The experimental results show that DeepIPC achieves the best performance even with fewer parameters. This establishes the effectiveness of our network architecture. %which is widely used in many driving simulator programs
	\item We introduce an innovative evaluation metric focused on driver interventions, a critical measure ensuring safety and practicality in deployment scenarios. This metric not only underscores our model's operational superiority but also demonstrates DeepIPC's efficiency and architectural elegance, achieving outstanding performance while maintaining a lean parameter profile, which underscores the model's design sophistication and practical applicability in real-world autonomous driving.
\end{itemize}

%%%%%%%%%%%%%%%%%%%%%%%%%%%%%%%%%%%%%%%%%%%%%%%%%%%%%%%%%%%%%%%%%%%%%

\section{Related Works}

In this section, we review some related works that focus on end-to-end autonomous driving. Then, we point out the key ideas which inspire our work and as an objective for comparative study.

\subsection{Perception-Action Coupling} \label{subs:perception_action}

Among various approaches in the field of autonomous driving, perception has always been the first stage as it is important to understand the surrounding area before planning and action. It can be achieved by performing various vision tasks such as semantic segmentation, depth estimation, and object detection \cite{matsuzaki}\cite{yubao}\cite{matsuzaki2}\cite{minami}. In the autonomous driving area, Hahner et al. proposed a segmentation model that is made specifically to deal with foggy conditions \cite{seg_fog}. Then, different work is proposed by Rajaram et al. \cite{obj_detect1} where a model called RefineNet is used to perform object detection. Besides completing a single vision task, the model can be pushed further to perform multiple vision tasks simultaneously to achieve a better scene understanding \cite{vis1}\cite{av_vision1}.

After achieving the ability to perceive the environment, a model also needs to leverage this ability to support the controller parts. In the field of end-to-end autonomous driving where perception and control are coupled together, better visual perception means better drivability as the controller gets better features directly from the perception module \cite{percep_control0}. The work in coupling perception and control modules has been done by Ishihara et al. \cite{highcommand1} where an end-to-end model is deployed to perform multiple vision tasks and predict navigational controls at the same time. Similar work is also proposed by Chitta et al. \cite{aim_mt} where a model called AIM-MT (auto-regressive image-based model with multi-task supervision) completes perception and control tasks simultaneously to drive a vehicle. It is disclosed that performing vision tasks can improve drivability as the controller receives better perception features.

Similar to Ishihara et al. \cite{highcommand1} and Chitta et al. \cite{aim_mt}, the perception parts of DeepIPC are guided by completing a vision task to provide better features. However, it only uses semantic segmentation as auxiliary supervision since the depth is considered as an input. Then, the controller is equipped with two decision-makers that predict waypoints and navigational controls to consider different aspects of driving. For a comparative study, we use AIM-MT as a baseline in justifying the performance of DeepIPC. The objective is to compare our model (that has a better data representation) with a model that is guided by extra supervision to produce better features for the controller.

\subsection{Multi-modal Fusion} \label{subs:rgbd_mod}
Processing one kind of data modality is not reliable for autonomous driving as it can be failed under certain conditions. Therefore, more heterogeneous data is needed to cover each other's weaknesses and produce more meaningful information through sensor fusion techniques \cite{rgbdvs}\cite{taro}. Some studies have been conducted in the field of sensor fusion for autonomous driving. To be more specific in the camera-LiDAR fusion, Prakash et al. \cite{transfuser} have proposed a model that processes RGB images and point clouds to perceive the environment and drive the vehicle. A certain transformer-based module called TransFuser \cite{transfuser2} is used to learn the relation between RGB images and point clouds to achieve better perception. Then, another work is proposed by Niesen and Unnikrishnan \cite{fuse4} where camera and radar are fused to achieve accurate 3D depth reconstruction in highway environments.

Mounting two different sensors can be another issue as more space, equipment, and extra budgets are needed. Therefore, using an equivalent sensor that is cheaper and can do a similar function may be preferable to tackle this problem. For example, a LiDAR can be replaced with a depth camera (merged with an RGB camera) to perceive the depth \cite{rgbd2}. In the use of RGBD image for autonomous driving, Huang et al. \cite{huang_model} demonstrated how RGB image and depth map can be fused and extracted from the early perception stage to provide better features for the controller. Besides pixel-to-pixel fusion, the depth map can be also projected to produce a BEV semantic map \cite{oskar_tiv}. Thus, the model can perceive from the top-view perspective for a better perception.

Together with AIM-MT \cite{aim_mt}, we deploy Huang et al.'s model \cite{huang_model} for a comparative study with the objective of comparing the performance of different sensor fusion strategies. Huang et al. fuse the information by processing RGB and depth at the early stage to extract a deeper relation on each pixel. Meanwhile, DeepIPC fuses the information by performing BEV semantic mapping to get the advantage of perceiving from a different perspective.

\subsection{Real-world Imitation Learning} \label{subs:imit_real}
In order to achieve end-to-end autonomous driving, one approach is to proceed with behavior cloning or imitation learning strategies which can be done easily in a supervised learning manner. By using the end-to-end imitation learning strategy, we can create a single deep learning model to imitate the behavior of an expert driver in manipulating navigational controls or effectors for handling complicated situations on the street \cite{sasaki}\cite{shen}. This can be derived from publicly available datasets or simulated with a simulator to enrich the model's driving experiences \cite{imit_sim_driving_data}\cite{imit_real_driving_data}. Therefore, the model will be able to perform human-like autonomous driving \cite{human_driving}.

Imitation learning has been widely used for real-world experiments. In the application to mobile robotics and autonomous vehicles, Cai et al. \cite{il_mro0} propose a vision-based model for driving a toy-size autonomous race car in a fixed circuit. This work shows how the imitation learning technique can be used to train a simple model to learn the mapping function between an RGB image as the input and navigational controls as the outputs. Not only simple models, but this technique is also applicable to multi-input multi-output models that process multiple data. Recent work by Chatty et al. \cite{il_mro1} demonstrates the use case of imitation learning for cognitive map building used for navigating a mobile robot. Then, Hoshino et al. \cite{il_mro2} also use the imitation-based end-to-end multi-task learning technique for motion planning and controlling a mobile robot in a challenging environment. Another similar work is proposed by Yan et al. \cite{il_fish} where an end-to-end model is used to control a robotic shark. These models are supported with multiple sensors and are used to control several end-effectors. Although they look promising, imitation learning sometimes causes an issue of generalization ability in new environments. % 

Following the success of these works in using imitation learning for complex multi-input multi-output models, we also use this approach to train DeepIPC for driving a robotic vehicle in real environments. Meanwhile, to overcome the generalization ability problem, we employ two control agents to manipulate the vehicle's end-effectors. As there are more decision-makers in its architecture, DeepIPC will be able to consider different aspects of drivability.

%%%%%%%%%%%%%%%%%%%%%%%%%%%%%%%%%%%%%%%%%%%%%%%%%%%%%%%%%%%%%%%%%%%%%

\section{Methodology}
%, 
In this section, we explain the model architecture and its improvement in detail. Then, we describe the dataset used for imitation learning that includes data for training, validation, and testing. We also describe the training setup and define several formulas used to supervise the model. Finally, we explain the evaluation setting including the metrics used to justify the model performance.

\subsection{Proposed Model} \label{subs:model}

\begin{figure*}[t]
	\begin{center}
		\includegraphics[width=\linewidth]{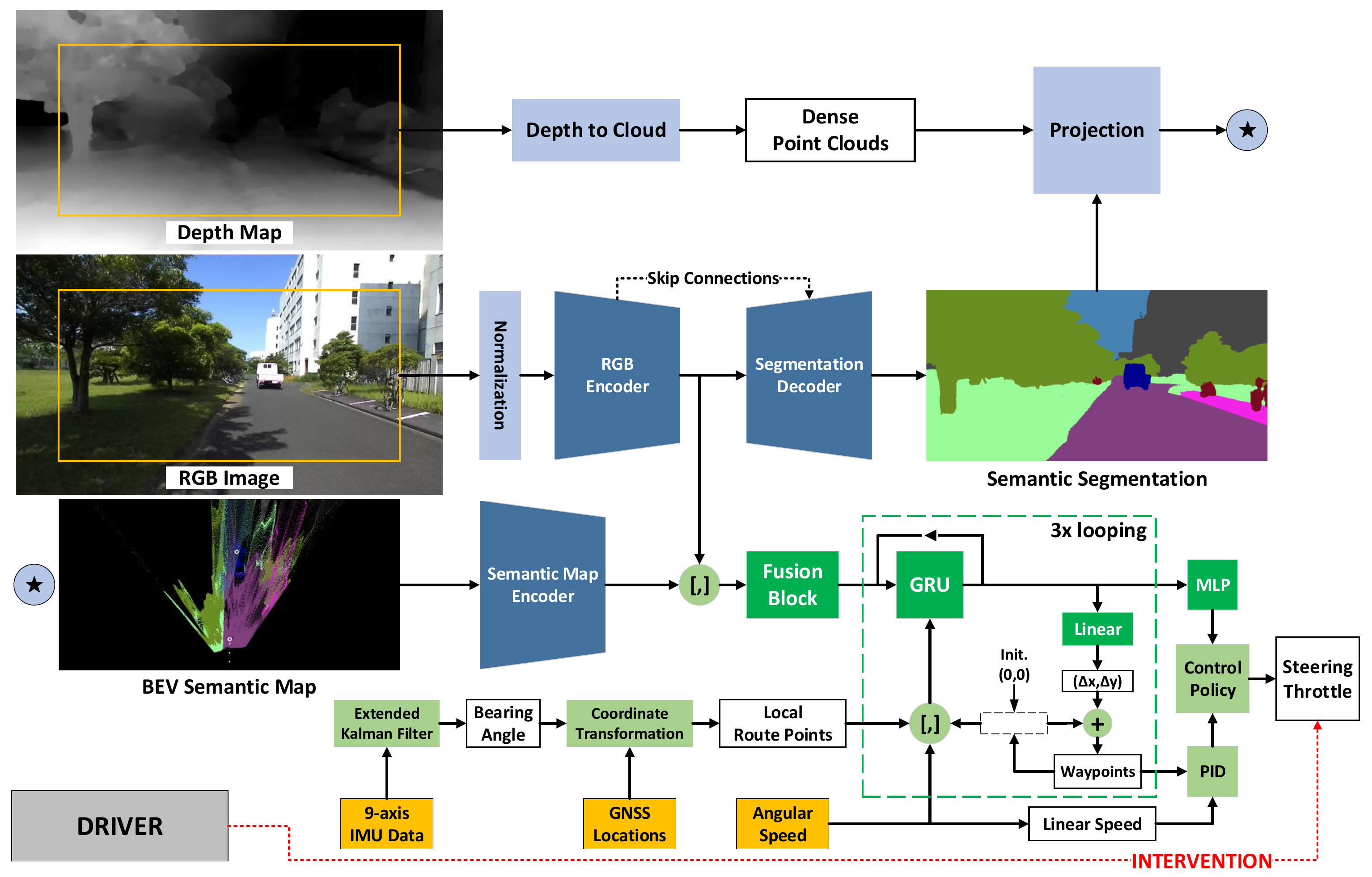}
		
	\end{center}
	%\vspace{-5mm}
	\caption{The architecture of DeepIPC. Blue blocks are parts of the perception module, while green blocks are parts of the controller module. Light-colored blocks are not trainable, while the darker ones are trainable. In the BEV semantic map, waypoints are denoted with white dots, while route points are denoted with white circles. Waypoints are points predicted by the DeepIPC (controller module) based on the features extracted by the perception module. These points are then translated into steering and throttle commands by two PID controllers to navigate the vehicle. Meanwhile, route points refer to points in global latitude-longitude coordinates that inform DeepIPC about the path for navigating the vehicle from the starting point to the destination. Additionally, route points can be generated with the assistance of applications such as Google Maps, providing a set of latitude-longitude coordinates that delineate the path from the starting location to the destination.}
	\label{fig:model}
	%\vspace{-3mm}
\end{figure*}

As mentioned in Section \ref{sec:intro}, the architecture of DeepIPC is based on our previous work \cite{oskar_tiv} that is composed of perception and controller. As shown in Fig. \ref{fig:model}, the perception phase begins with semantic segmentation on RGB image with a standard encoder-decoder network enhanced with several skip connections \cite{unetblock}\cite{araki}\cite{masuzawa}. The RGB encoder is made of a pretrained Efficient Net B3 \cite{effnet} while the decoder is composed of multiple convolution blocks where each block consists of ($2\times$($3\times3$ convolution $+$ batch normalization \cite{batchnorm} $+$ ReLU \cite{relu}) $+$ bilinear interpolation) and a pointwise $1\times1$ convolution followed with sigmoid activation. Furthermore, we generate point clouds from the depth map and make projections with the predicted segmentation map to obtain a BEV semantic map with a coverage area of 24 meters to the front, left, and right from the vehicle location. Thus, the vehicle is always positioned at the bottom center of the BEV semantic map. Then, the BEV semantic map is encoded by an Efficient Net B1 encoder \cite{effnet} to obtain its features. With this configuration, DeepIPC has both front and top perspectives to perceive the surrounding area.

In the controller module, both RGB and BEV semantic features are processed by a fusion block module which is composed of pointwise $(1\times1)$ convolution, global average pooling, and linear layer. This module is responsible for learning the relation between features from the front and top-view perspectives and resulting in more compact latent features. Then, a gated recurrent unit (GRU) \cite{gru} is used to decode the latent features based on the measurement of the left and right wheel's angular speeds, predicted waypoints, and two route points that have been transformed into local BEV coordinates. The decoded features are processed further by a linear layer to predict $\Delta x$ and $\Delta y$. Thus, the next waypoint coordinate $(x_{i+1}, y_{i+1})$ can be calculated with (\ref{eq:wp}).

%\vspace{-3mm}
\begin{equation} \label{eq:wp}
	x_{i+1}, y_{i+1} = (x_{i}+\Delta x), (y_{i}+\Delta y)
\end{equation}

During the training process, waypoints represent the vehicle's locations at time $t+1$, $t+2$, and $t+3$ seconds in the future, projected onto a Cartesian coordinate system where the position (0,0) corresponds to the vehicle's current location at time $t$. In essence, waypoints indicate the expected future positions of the vehicle based on the current observations. To be noted, the waypoints prediction process is looped over three times as there are three waypoints to be predicted. In the first loop, the waypoint is initialized with the vehicle position in the local BEV coordinate which is always at (0,0). In the end, the waypoints are translated into a set of navigational controls (steering and throttle) by two PID controllers in which their $Kp, Ki, Kd$ parameters are tuned empirically. The final features used to predict the last waypoint are also processed by a multi-layer perceptron (MLP) block to estimate the navigational controls directly. We have two main reasons for enabling DeepIPC to predict three waypoints:
\begin{itemize}
	\item The first and second waypoints are utilized by PID agent for lateral and longitudinal control of the vehicle. After predicting the second waypoint, the latent space is adjusted using the latest waypoint (the second one), ensuring that the MLP agent receives a comparable level of information abstraction as the PID agent. This adjustment results in the generation of the third waypoint, which the PID agent do not utilize.
	\item Incorporating the third waypoint into the loss calculation enhances the training signal for DeepIPC, enabling more accurate predictions of the vehicle's future positions. Nonetheless, we restrict the number of predicted waypoints to three to ensure that the MLP agent processes information congruently with the PID agent.
\end{itemize}

The final action that actually drives the vehicle is made by a control policy that combines both PID and MLP controls as shown in Algorithm \ref{alg:control}. To ensure smooth control, we set a minimum threshold of 0.1 for both steering and throttle commands issued by an agent. This translates to requiring the predicted control values, exceeding or equaling 0.1, before any movement occurs. This approach also enables one agent to fully takeover control from another, enhancing overall maneuverability. Keep in mind that the takeover is initiated when the agent's output falls below a critical confidence level of 0.1. To ensure the correctness of these takeover actions, DeepIPC employs a robust validation mechanism, incorporating cross-referencing of agent's decisions and a dynamic evaluation protocol that continuously monitors environmental interactions and agent performance. This protocol facilitates a swift and reliable transition between agents, maintaining system integrity and operational fluency even under challenging or unpredictable conditions. Through systematic testing and validation outlined in our experimental results discussed in Section \ref{sec4}, we demonstrate the efficacy and precision of our agent takeover strategy, reinforcing DeepIPC's resilience and adaptability in diverse operational contexts.

\begin{algorithm}[!t] 
	\SetAlgoLined
	%Input:\\
	%$Wp_{\{1,2\}}$, $\mathbf{MLP}_{\{ST,TH\}}$, $\omega_{\{l,r\}}$\\
	%\dotfill
	%$ $\\
	%PID's steering and throttle:\\
	$\Theta = \frac{Wp_1+Wp_2}{2}$\\
	$\theta = \tan^{-1}\big(\frac{\Theta[1]}{\Theta[0]}\big)$\\ 
	$\gamma = 1.75\times||Wp_1 - Wp_2||_F$\\
	$\nu = \frac{(\omega_l + \omega_r)}{2} \times r$\\
	\dotfill
	$ $\\
	$\mathbf{PID}_{ST} = \mathbf{PID}^{Lat}(\theta-90)$\\ 
	$\mathbf{PID}_{TH} = \mathbf{PID}^{Lon}(\gamma-\nu)$\\
	\dotfill
	$ $\\
	%Final decision:\\
	\uIf{$\mathbf{MLP}_{TH} \geq 0.1$ and $\mathbf{PID}_{TH} \geq 0.1$}{
		\uIf{$|\mathbf{MLP}_{ST}| \geq 0.1$ and $|\mathbf{PID}_{ST}| < 0.1$}{
			steering $= \mathbf{MLP}_{ST}$\\
		}
		%\uElseIf
		\uIf{$|\mathbf{MLP}_{ST}| < 0.1$ and $|\mathbf{PID}_{ST}| \geq 0.1$}{
			steering $= \mathbf{PID}_{ST}$\\
		}
		\uElse{
			steering $= \beta_{00} \mathbf{MLP}_{ST} + \beta_{10} \mathbf{PID}_{ST}$\\
		}
		throttle $= \beta_{01} \mathbf{MLP}_{TH} + \beta_{11} \mathbf{PID}_{TH}$\\
	}
	\uElseIf{$\mathbf{MLP}_{TH} \geq 0.1$ and $\mathbf{PID}_{TH} < 0.1$}{
		steering $= \mathbf{MLP}_{ST}$\\
		throttle $= \mathbf{MLP}_{TH}$\\
	}
	\uElseIf{$\mathbf{MLP}_{TH} < 0.1$ and $\mathbf{PID}_{TH} \geq 0.1$}{
		steering $= \mathbf{PID}_{ST}$\\
		throttle $= \mathbf{PID}_{TH}$\\
	}
	\uElse{
		steering $= 0$\\
		throttle $= 0$\\
	}
	$ $\\
	\dotfill\\
	\caption{Control Policy}
	\label{alg:control}
	\begin{tablenotes}
		\small
		%\item --------------------------------------------------------------------------------
		\item $Wp_{\{1,2\}}$: first and second predicted waypoints
		\item $\mathbf{MLP}_{\{ST,TH\}}$: steering and throttle estimated by MLP %agent
		\item $\omega_{\{l,r\}}$: angular speed of left/right wheel% measured with rotary encoder
		\item $r$: vehicle's wheel radius, 0.15 m 
		\item $\Theta$: aim point, a middle point between $Wp_1$ and $Wp_2$
		\item $\theta$: heading angle derived from the aim point $\Theta$
		\item $\gamma$: desired speed, Frobenius norm of $Wp_1$ and $Wp_2$ %1.75 $\times$ 
		\item $\nu$: linear speed, average of $\omega_l$ and $\omega_r$ multiplied by $r$
		\item $\beta\in\{0,...,1\}^{2\times2}$ is a set of control weights:
		\item $\beta_{00} = \frac{\alpha_2}{\alpha_2+\alpha_1}$; $\beta_{10} = 1 - \beta_{00}$; $\beta_{01} = \frac{\alpha_3}{\alpha_3+\alpha_1}$; $\beta_{11} = 1 - \beta_{01}$
		\item where $\alpha_1, \alpha_2, \alpha_3$ are loss weights computed by MGN
		\item algorithm \cite{mgn} (see Subsection \ref{subs:train} for more details)
	\end{tablenotes}
\end{algorithm}

\subsection{Model Improvement} \label{subs:modif}

Different from our previous work \cite{oskar_tiv}, the model is modified to improve its performance and deal with real-world implementation issues. First, as the input to the perception module, we consider a wider ROI of $H \times W = 512 \times 1024$ at the center of the RGBD image. Then, they are resized to $H \times W = 256 \times 512$ to reduce the computational load. With a wider input resolution, the model is expected to have a better scene understanding capability. Second, as the input to the controller module, we feed the left and right wheel's angular speeds. This information is expected to be helpful, especially during turning as the angular speed will be different on each wheel. Third, two route points are given instead of one route point at a time. Relying only on one route point is very risky due to the possibility of sensor inaccuracies that affect the global-to-local coordinate transformation. Although we have used RTK-GNSS and achieved 1cm positional accuracy, the coordinate transformation can be still disturbed by the inaccuracy of the IMU sensor. If the route point is mislocated, the model will likely fail to predict waypoints and navigational controls correctly. Besides that, giving two route points will give the model a better intuition in deciding whether the vehicle should drive straight or turn depending on the location of the route points. Fourth, we also modify the control policy by allowing an agent to take the steering control completely over the other agent for better maneuverability.

In this research, we feed the model with data from the GNSS receiver and 9-axis IMU sensor to measure several quantities needed to perform global-to-local coordinate transformation precisely. To get the local BEV coordinate for each route point $i$, the relative distance $\Delta x_{i}$ and $\Delta y_{i}$ between vehicle location $Ro$ and route point location $Rp_i$ must be known. The distance can be estimated from the global longitude-latitude with (\ref{eq:lon_meter}) and (\ref{eq:lat_meter}).

%\vspace{-7mm}
\begin{equation} \label{eq:lon_meter}
	\Delta x_{i} = (Rp^{Lon}_i - Ro^{Lon}) \times \frac{\mathcal{C}_e \times \cos(Ro^{Lat})}{360},
\end{equation}
%\vspace{-4mm}
\begin{equation} \label{eq:lat_meter}
	\Delta y_{i} = (Rp^{Lat}_i - Ro^{Lat}) \times \frac{\mathcal{C}_m}{360},
\end{equation}

\begin{figure}[!t]
	\begin{center}
		\includegraphics[width=\linewidth]{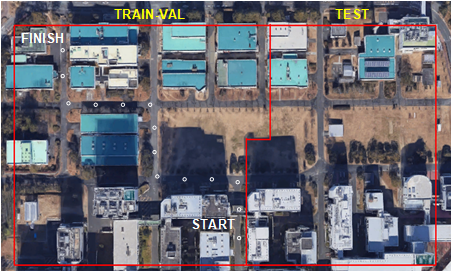}
		
	\end{center}
	%\vspace{-4mm}
	\caption{The experiment area. White hollow circles represent a route that consists of start, finish, and a set of route points. (\href{https://goo.gl/maps/9rXobdhP3VYdjXn48}{https://goo.gl/maps/9rXobdhP3VYdjXn48})}
	\label{fig:tut_map}
	%\vspace{-3mm}
\end{figure}

\noindent where $\mathcal{C}_e$ and $\mathcal{C}_m$ are earth's equatorial and meridional circumferences which are around 40,075 and 40,008 kilometers, respectively. Then, the route point coordinates $Rp^{(x,y)}_i$ can be obtained by applying a rotation matrix as in (\ref{eq:bev_transform}).

%\vspace{-5mm}
\begin{equation} \label{eq:bev_transform}
	\begin{bmatrix}
		Rp^{x}_i \\
		Rp^{y}_i 
	\end{bmatrix} = 
	\begin{bmatrix}
		\cos(\theta_{ro}) & -\sin(\theta_{ro}) \\
		\sin(\theta_{ro}) & \cos(\theta_{ro})
	\end{bmatrix}^T
	\begin{bmatrix}
		\Delta x_{i}\\
		\Delta y_{i}
	\end{bmatrix},
\end{equation}

\noindent where $\theta_{ro}$ is the vehicle's absolute orientation to the north pole (bearing angle). It is estimated by a 9-axis IMU sensor's built-in function based on Kalman filtering on 3-axial acceleration, angular speed, and magnetic field. The global-to-local route points transformation may not be so accurate due to sensor inaccuracy or noisy measurement. Hence, the model is forced implicitly to learn how to compensate for this issue by mimicking expert behavior for estimating the waypoints and navigational controls. 
%We use the extended Kalman filter to estimate $\theta_{ro}$ based on the measurement of 3-axial acceleration, angular speed, and magnetic field retrieved from a 9-axis IMU sensor.

%(\ref{eq:bearing})
%\begin{equation} \label{eq:bearing}
%	\theta_{ro} = \mathbf{EKF}(\alpha,\omega,m)
%\end{equation}

%where $\alpha, \omega, m$ are acceleration, angular speed, and magnetic field on three different axis $x,y,z$ measured by a 9-axis IMU.

\subsection{Dataset}\label{subs:dataset}

\begin{figure}[!t]
	\begin{center}
		\includegraphics[width=0.985\linewidth]{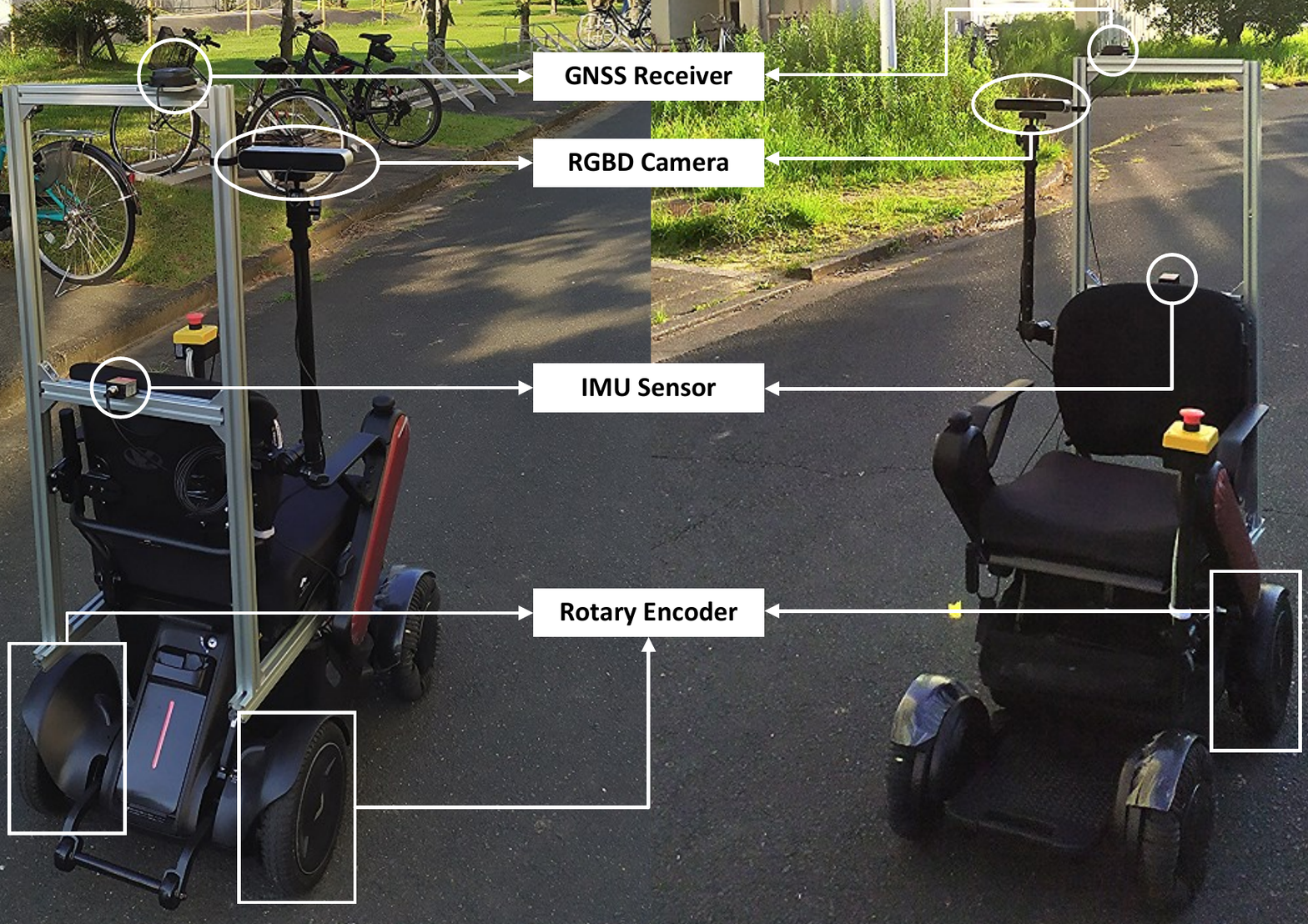}
		
	\end{center}
	%\vspace{-4mm}
	\caption{Sensor placement on a robotic vehicle. The rotary encoder is mounted inside each rear wheel.} %  
	\label{fig:vehicle}
	%\vspace{-4mm}
\end{figure}

% 
% \footnote{\url{https://goo.gl/maps/9rXobdhP3VYdjXn48}}
% 
In imitation learning and behavior cloning, a considerable amount of expert driving records is needed for training and validation (train-val) \cite{imitation2}\cite{garcia}\cite{hidehito}\cite{uzawa}. To create the dataset, we drive the vehicle at a speed of 1.25 m/s in a certain area inside Toyohashi University of Technology, Japan. As shown in Fig. \ref{fig:tut_map}, the left region is used for the train-val, and the right region is used for the test. We consider two experiment conditions which are noon and evening. For each condition, we record the driving data one time for the train-val and three times for the test. There are 12 routes in the train-val region and 6 routes in the test region. Each route is composed of several route points with a gap of 12 meters between each other. The route point will be changed once the vehicle is 4 meters close. In completing the point-to-point navigation task, the model must follow the route points in driving the vehicle. The observation is recorded at 4 Hz and composed of an RGBD image, GNSS location, 9-axis IMU measurement, the wheel's angular speed, and the level of steering and throttle. The devices used to retrieve the data are mentioned in Table \ref{tab:data_info}, while their placement can be seen in Fig. \ref{fig:vehicle}.  

\begin{table}[t] %[hb]
	\caption{Dataset Information}
	%\vspace{-5mm}
	\begin{center}
		\resizebox{\linewidth}{!}{%
			%\begin{tabular*}{\textwidth}{@{\extracolsep{\stretch{1}}}*{7}{r}@{}}
			\begin{tabular}{p{0.275\linewidth}p{0.65\linewidth}}
				% \begin{tabular}{cc}
					\toprule
					% \hdashline\noalign{\vskip 0.75ex}
					
					Conditions & Noon and evening\\
					\midrule
					Total routes & 12 (train-val)\\
					& 6 (test)\\
					\midrule
					%Train frames & 4831 (noon), 5320 (evening), 10151 (total)\\
					%Val frames & 4863 (noon), 4816 (evening), 9679 (total)\\
					%Test frames & 9510 (noon), 9465 (evening), 18975 (total)\\
					$\mathcal{N}$ samples* & 10151 (train) and 9679 (val)\\
					& 18975 (test)\\
					\midrule
					Devices & WHILL C2 vehicle\\
					& Stereolabs Zed RGBD camera\\
					& U-blox Zed-F9P GNSS receiver\\
					& Witmotion HWT905 IMU sensor\\
					& WHILL's rotary encoder\\
					
					% \hdashline\noalign{\vskip 0.75ex}
					\midrule
					Object classes & None, road, sidewalk, building, wall, fence, pole, traffic light, traffic sign, vegetation, terrain, sky, person, rider, car, truck, bus, train, motorcycle, bicycle\\
					% \hdashline\noalign{\vskip 0.75ex}
					
					\bottomrule                             
			\end{tabular}
		}
	\end{center}
	\label{tab:data_info}
	%\vspace{-2mm}
	\begin{tablenotes}\small
		\item *$\mathcal{N}$ samples is the number of observation sets. Each set consists of an RGBD image, GNSS location, 9-axis IMU measurement, wheel's angular speed, and the level of steering and throttle.
	\end{tablenotes}
	%\vspace{-1mm}
\end{table}

As mentioned in Subsection \ref{subs:model}, DeepIPC predicts waypoints, navigational controls, and semantic segmentation maps. As for waypoints ground truth, we leverage the vehicle's trajectory where the vehicle's location in one second, two seconds, and three seconds in the future are considered as the waypoints to be predicted. Meanwhile, navigational controls ground truth can be obtained from the record of steering and throttle levels. To avoid time-consuming manual annotation, we use SegFormer \cite{segformer} pre-trained on the Cityscapes dataset \cite{cityscapes} to provide pseudo-labels by performing semantic segmentation on all RGB images in twenty different classes as mentioned in Table \ref{tab:data_info}. SegFormer is chosen as it is one of the state-of-the-art models that has excellent performance in the image segmentation task.

\subsection{Training}\label{subs:train}
With using the multi-task learning paradigm, DeepIPC can be supervised by a combination of several weighted loss functions as in (\ref{eq:mtlloss}).

%train_sunny = 4831
%train_sunset = 5320
%train_total = 10151

%val_sunny = 4863
%val_sunset = 4816
%val_total = 9679

%test_sunny123 = 3230 + 3164 + 3116 = 9510
%test_sunset123 = 3203 + 3136 + 3126 = 9465
%test_total123 = 18975

%\vspace{-6mm}
\begin{equation} \label{eq:mtlloss}
	\mathcal{L}_{MTL} = \alpha_0\mathcal{L}_{SEG} + \alpha_1\mathcal{L}_{WP} + \alpha_2\mathcal{L}_{ST} + \alpha_3\mathcal{L}_{TH},
\end{equation}

\noindent where $\alpha_{0,1,2,3}$ are loss weights tuned adaptively by an algorithm called modified gradient normalization (MGN) \cite{mgn} to ensure that all tasks can be learned at the same pace. To learn semantic segmentation, we use a combination of pixel-wise cross entropy and dice loss as in (\ref{eq:bcedice}). 

%\vspace{-6mm}
\begin{equation} \label{eq:bcedice}
	\begin{split}
		\mathcal{L}_{SEG} & = \bigg( \frac{1}{N} \sum_{i = 1}^{N} y_i log(\hat{y}_i) + (1-y_i) log(1-\hat{y}_i) \bigg)\\
		& + \bigg( 1 - \frac{2|\hat{y} \cap y|}{|\hat{y}| + |y|} \bigg),
	\end{split}
\end{equation}

\noindent where $N$ is the total elements at the last layer of the segmentation decoder, while $y_i$ and $\hat{y}_i$ are ground truth and prediction of element $i$. Then, we use L1 loss to supervise waypoints prediction as in (\ref{eq:l1loss1}).

%\vspace{-2mm}
\begin{equation} \label{eq:l1loss1}
	\mathcal{L}_{WP} = \frac{1}{M} \sum_{i = 1}^{M} |\hat{y}_i - y_i|,
\end{equation}

\noindent where $M$ is equal to 6 as there are three predicted waypoints that have x,y coordinates for each. Similarly, we use L1 loss to supervise steering and throttle estimation as in (\ref{eq:l1loss2}). However, averaging is not necessary as there is only one element for each output.% (). 

%\vspace{-3mm}
\begin{equation} \label{eq:l1loss2}
	\mathcal{L}_{\{ST,TH\}} = |\hat{y} - y|
\end{equation}

\begin{table*}[t] %[hb]
	\caption{Model Specification}
	%\vspace{-5mm}
	\begin{center}
		\resizebox{\textwidth}{!}{%
			%\begin{tabular*}{\textwidth}{@{\extracolsep{\stretch{1}}}*{7}{r}@{}}
			% \begin{tabular}{|p{0.2\linewidth}|p{0.2\linewidth}|p{0.1\linewidth}|p{0.4\linewidth}|}
				\begin{tabular}{ccccc}
					%\toprule
					%& $z_{6}$ & $z_{8}$ & $z_{9}$ & $z_{11}$ & $z_{13}$ & $z_{14}$ \\
					%\midrule
					% \hline
					\toprule
					% {\bf Model} & {\bf Total Parameters$\downarrow$} & {\bf GPU Load (MB)$\downarrow$} & {\bf Input/Sensor}\\
					Model & Total Parameters$\downarrow$ & Model Size $\downarrow$ & Input/Sensor & Output\\
					\toprule
					% \hline
					Huang et al. \cite{huang_model} & 74953258 & 300.196 MB & RGBD, High-level commands & Segmentation, Steering, Throttle\\
					% \hline
					AIM-MT \cite{aim_mt} & 27967063 & 112.078 MB & RGB, GNSS, 9-axis IMU, Rotary encoder & Segmentation, Depth, Waypoints, Steering, Throttle\\
					% \hline
					% \hdashline\noalign{\vskip 0.75ex}
					DeepIPC & 20983128 & 84.972 MB & RGBD, GNSS, 9-axis IMU, Rotary encoder & Segmentation, BEV Semantic, Waypoints, Steering, Throttle\\
					% \hline
					\bottomrule                             
				\end{tabular}
			}
	\end{center}
	%\vspace{-1mm}
	\label{tab:model_compare}
	\begin{tablenotes}\small
		%\item *These outputs are not supervised with a certain loss function during training process.
		%
		\item AIM-MT \cite{aim_mt} is implemented based on the codes shared in the author's repository at \href{https://github.com/autonomousvision/neat}{https://github.com/autonomousvision/neat}. Meanwhile, Huang et al.'s model \cite{huang_model} is implemented based on the explanation written in the paper. All models are deployed on a laptop powered with NVIDIA GTX 1650 GPU in performing real-world autonomous driving. As the models can run smoothly during evaluation, we believe that calculating their inference speeds is not necessary. However, we assume that a smaller model with less number of parameters is preferred as it does not consume large computation power. %DeepIPC is considered to be the smallest model as it has the lowest number of trainable parameters.
	\end{tablenotes}
	%\vspace{-2mm}
\end{table*}

The model is implemented with PyTorch deep learning framework \cite{torch} and trained on NVIDIA RTX 3090 by Adam optimizer \cite{optim_adam} with a decoupled weight decay \cite{adamw} of 0.001 and a batch size of 8. The initial learning rate is set to 0.0001 and divided by 2 if validation $\mathcal{L}_{MTL}$ is not dropping in 5 epochs. To avoid unnecessary computation, the training is stopped if there is no improvement in 30 epochs.

\subsection{Evaluation and Scoring} \label{subs:eval}
DeepIPC is evaluated under two conditions with varying cloud intensity with two different tests namely offline and online tests. For each condition, the final score is obtained by averaging the scores from three experimental results. In the offline test, the model is deployed to predict driving records. Then, its performance on each task is calculated by a specific metric function. To evaluate waypoints and navigational controls, we use mean absolute error (MAE) or L1 loss as in (\ref{eq:l1loss1}) and (\ref{eq:l1loss2}). Meanwhile, we compute intersection over union (IoU) as in (\ref{eq:iou}) for evaluating the segmentation performance.

%\vspace{-2mm}
\begin{equation} \label{eq:iou}
	IoU_{SEG} = \frac{|\hat{y} \cap y|}{|\hat{y} \cup y|}
\end{equation}

We define the best model by the lowest total metric (TM) score as formulated with (\ref{eq:tm}) that combines segmentation IoU and controls estimation MAE. Depth and waypoints MAE are excluded since not every model has these outputs.

%\vspace{-5mm}
\begin{equation} \label{eq:tm}
	TM = (1-IoU_{SEG}) + MAE_{ST} + MAE_{TH}
\end{equation}

\begin{algorithm}[!t] 
	\SetAlgoLined
	%Input: $Rp^x_{\{1,2\}}$\\
	%$ $\\
	\uIf{$Rp^x_1 \leq -4m$ or $Rp^x_2 \leq -8m$}{
		command = turn left
	}
	\uElseIf{$Rp^x_1 \geq 4m$ or $Rp^x_2 \geq 8m$}{
		command = turn right
	}
	\uElse{
		command = go straight
	}
	
	\dotfill
	\caption{Route points to Commands}
	\label{alg:nav_cmd}
	\begin{tablenotes}
		\small
		%\item --------------------------------------------------------------------------------
		\item $Rp^x_{\{1,2\}}$: the route point's local $x$ position %in local BEV coordinate
	\end{tablenotes}
\end{algorithm}

In the online test, the model is deployed to drive a robotic vehicle by following a set of routes. Unlike in our previous work \cite{oskar_tiv}, the vehicle is prevented from colliding with other objects as it can cause unnecessary damage. Thus, we determine the drivability score by counting the number and time of interventions needed to prevent collisions.

% 
% as mentioned in Subsection \ref{subs:rgbd_mod}
%
In addition, we conduct a comparative study with some recent models to get a clearer performance justification. Table \ref{tab:model_compare} shows the specification of the models evaluated in this study where our DeepIPC is considered to be the smallest model as it has the lowest number of parameters. We evaluate a model proposed by Huang et al. \cite{huang_model} that takes RGB images and depth maps but with a different fusion strategy. This model uses high-level commands in selecting a command-specific controller. Hence, we generate these commands automatically based on the route point position in the local coordinate using a certain rule as described in Algorithm \ref{alg:nav_cmd}. We also evaluate AIM-MT \cite{aim_mt} which only takes RGB images and predicts multiple vision tasks for extra supervision. By performing more vision tasks, the perception module can provide better features for the controller. For a fair comparison, we modify both models to process the same information as provided to DeepIPC. %The model specification can be seen in Table \ref{tab:model_compare}. 

In our comparative network architecture analysis, while the AIM-MT \cite{aim_mt} shows commendable integration of multiple vision tasks to augment controller input, it is constrained by its reliance on RGB data alone, which may limit performance under variable lighting conditions or complex environments. In contrast, DeepIPC leverages RGBD inputs, providing a richer, more robust data source for perception-action coupling. On the other hand, Huang et al.'s \cite{huang_model} approach to early-stage fusion of RGB and depth data offers a novel perspective on multi-modal integration, yet it may face challenges in disentangling conflicting features. Our DeepIPC model seeks to balance these aspects, utilizing BEV mapping for enhanced spatial awareness and context, a critical advantage in dynamic and unpredictable real-world settings.

%%%%%%%%%%%%%%%%%%%%%%%%%%%%%%%%%%%%%%%%%%%%%%%%%%%%%%%%%%%%%%%%%%%%%%%%%%%

\begin{table*}[t]
	\caption{Multi-task Performance Score}
	%\vspace{-5mm}
	\begin{center}
		\resizebox{\linewidth}{!}{%
			\begin{tabular}{ccccccccc}%cc
				\toprule
				Condition&Model&Total Metric$\downarrow$&$IoU_{SEG}\uparrow$&$MAE_{DE}\downarrow$&$MAE_{WP}\downarrow$&$MAE_{ST}\downarrow$&$MAE_{TH}\downarrow$&Latency$\downarrow$\\
				\toprule %\midrule
				%%%%%%%%%%%
				&Huang et al. \cite{huang_model} &0.4778 $\pm$0.0281       &0.8300 &-      &-      &0.2422 &0.0484 &0.0190\\%&&\\
				Noon &AIM-MT \cite{aim_mt}       &0.2932 $\pm$0.0300       &0.8863 &0.0593 &0.0983 &0.1734 &{\bf 0.0061} &0.0172\\%&&\\
				&{\bf DeepIPC}                      &{\bf 0.2807 $\pm$0.0335} &{\bf 0.8899} &-      &{\bf 0.0683} &{\bf 0.1632} &0.0074 &{\bf 0.0090}\\%&&\\
				\midrule
				%%%%%%%%%%%
				&Huang et al. \cite{huang_model} &0.4875 $\pm$0.0453       &0.7952 &-      &-      &0.2384 &0.0443 &0.0182\\%&&\\
				Evening &AIM-MT \cite{aim_mt}    &0.3088 $\pm$0.0346       &0.8578 &0.0669 &0.0931 &0.1639 &{\bf 0.0026} &0.0156\\%&&\\
				&{\bf DeepIPC}  				     &{\bf 0.3030 $\pm$0.0369} &{\bf 0.8623} &-      &{\bf 0.0645} &{\bf 0.1611} &0.0041 &{\bf 0.0097}\\%&&\\
				\bottomrule
			\end{tabular}
		}
	\end{center}
	\label{tab:mtl_result}
	%\vspace{-2mm}
	\begin{tablenotes}\small
		\item The best performance is defined by the lowest total metric score as formulated with (\ref{eq:tm}). $IoU_{SEG}$: IoU score of semantic segmentation. $MAE_{DE}$: mean absolute error of normalized depth estimation. $MAE_{WP}$: mean absolute error of waypoints prediction. $MAE_{ST}$: mean absolute error of steering estimation. $MAE_{TH}$: mean absolute error of throttle estimation. Latency: the time (in seconds) required by the model to process one set of observation data. To ensure a fair latency calculation, we run each model independently on the same PC with NVIDIA RTX 3090.% The depth output is normalized from 0.2 to 40 meters to 0 to 1
	\end{tablenotes}
	%\vspace{-2mm}
\end{table*}

\section{Result and Discussion} \label{sec4}
\begin{figure*}[t]
	\begin{center}
		\includegraphics[width=\linewidth]{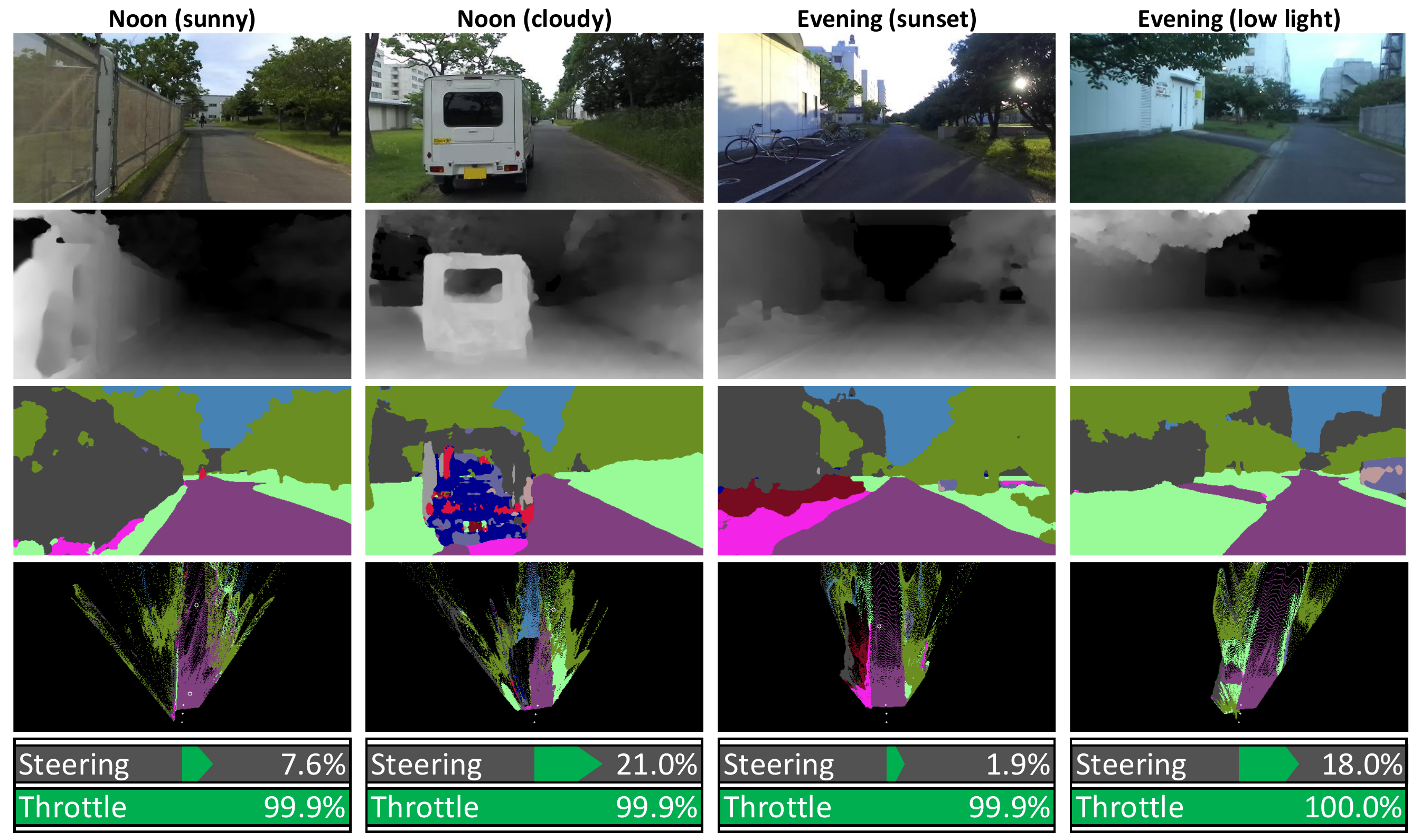}
		
	\end{center}
	%\vspace{-5mm}
	\caption{Driving footage. See the driving video (playback speed 5$\times$) at \href{https://youtu.be/AiKotQ-lAzw}{https://youtu.be/AiKotQ-lAzw} for more details, including failure cases where we intervene in the model to avoid collisions. Sunny noon: DeepIPC makes a small steering adjustment to the right as the vehicle is too close to the terrain. Cloudy noon: Although DeepIPC cannot segment the car properly, it can avoid collision as it knows that the left side is occupied. Sunset evening: DeepIPC makes a small steering adjustment to keep on its lane. Low light evening: We intervene in DeepIPC to avoid driving off-road on the vegetation as it keeps the throttle maximum and fails to make a right turn.}
	\label{fig:driving_rec}
	%\begin{tablenotes}\small
	%	\item Sunny noon: DeepIPC makes a small steering adjustment to the right as the vehicle is too close to the terrain.
	%	\item Cloudy noon: Although DeepIPC cannot segment the car properly, it can avoid collision as it knows that the left side is occupied.
	%	\item Sunset evening: DeepIPC makes a small steering adjustment to keep on its lane.
	%	\item Low light evening: DeepIPC makes a right turn following the route points.
	%\end{tablenotes}
	%\vspace{-3mm}
\end{figure*}

To justify its performance, DeepIPC is evaluated using two different methods namely the offline test and the online test. The results can be seen in Table \ref{tab:mtl_result} and Table \ref{tab:drive_result} respectively. Meanwhile, the qualitative results are shown in Fig. \ref{fig:driving_rec}.

\subsection{Offline Test} 
The offline test is used to evaluate the model's performance in handling multiple perception and control tasks simultaneously. All models are deployed to predict driving records and evaluated with multi-task and task-wise scoring. The test dataset is recorded three times in a completely different area from the train-val dataset. Each record is taken on different days to vary the situation and cloud intensity.

Table \ref{tab:mtl_result} shows that DeepIPC achieves the best performance by having the lowest total metric score in all conditions. Moreover, it achieves the fastest inference speed (lowest latency) as it has the lowest number of parameters, yielding a very low computational load compared to the other models. However, all models including DeepIPC have performance degradation in the evening. This means that doing inference in the low light condition is harder than in the normal condition. Specifically, in the segmentation task, DeepIPC has a higher IoU than AIM-MT even though it does not perform depth estimation for extra supervision that can enhance the RGB encoder. Thanks to the end-to-end learning strategy where the segmentation prediction can be processed further through the encoding and decoding process of the BEV semantic map. Therefore, the segmentation decoder receives a more useful gradient signal to tune the network weights properly. Meanwhile, Huang et al.'s model has the worst segmentation performance caused by conflicting features from fusing RGB images and depth maps from the early perception stage.

In the waypoints prediction task, DeepIPC has a lower MAE compared to AIM-MT. Thanks to the BEV semantic features, DeepIPC can distinguish free and occupied areas easily from the top-view perspective. Thus, it can properly estimate the waypoints which are also laid in BEV space. Although AIM-MT predicts four waypoints and DeepIPC only predicts three waypoints, it is still considered a fair comparison because the MAE formula averages the error across all predictions. The reason the AIM-MT predicts four waypoints is to let its controller module have more learning experiences in estimating the waypoints correctly. However, DeepIPC still performs better as its controller module gets boosted by BEV semantic features and fed with angular speed measurement which enhances its intuition. This result is in line with the result in our previous work \cite{oskar_tiv} where the model that perceives in BEV perspective (by using depth projection or LiDAR) is better at estimating the waypoints than the model that perceives in front-view perspective only.

\begin{table}[!t]
	\caption{Drivability Score}
	%\vspace{-6mm}
	\begin{center}
		\resizebox{\linewidth}{!}{%
			\begin{tabular}{cccc}%cc
				\toprule
				\multirow{2}{*}{Condition} & \multirow{2}{*}{Model} & \multicolumn{2}{c}{Intervention$\downarrow$}\\
				& & Count & Time (secs)\\
				\toprule %\midrule
				%%%%%%%%%%%
				&Huang et al. \cite{huang_model}       &1.8889 $\pm$0.4157       & 5.6039 $\pm$1.7272\\%&&\\
				Noon &AIM-MT \cite{aim_mt}             &2.2778 $\pm$0.3425       & 4.2161 $\pm$0.8380\\%&&\\
				&{\bf DeepIPC}                            &{\bf 1.1111 $\pm$0.3928} &{\bf 2.3092 $\pm$0.9841}\\%&&\\
				\midrule
				%%%%%%%%%%%
				&{\bf Huang et al. \cite{huang_model}} &{\bf 1.6111 $\pm$0.2079} & 4.5532 $\pm$0.2160\\%&&\\
				Evening &AIM-MT \cite{aim_mt}          &2.6667 $\pm$0.1361       & 4.6736 $\pm$0.4293\\%&&\\
				&{\bf DeepIPC}  				           &1.8889 $\pm$0.3928 	     &{\bf 4.2286 $\pm$0.6102}\\%&&\\
				\bottomrule
			\end{tabular}
		}
	\end{center}
	\label{tab:drive_result} 
	%\vspace{-3mm}
	\begin{tablenotes}\small
		\item The best drivability is defined by the lowest intervention count and intervention time. In the online test, there is no need to measure the inference speed as we limit the observation sampling to 4 Hz (the same configuration as the data gathering process used for training and validation) to perform a fair evaluation for each model.
	\end{tablenotes}
	\vspace{-1mm}
\end{table}

In the navigational controls estimation task, DeepIPC also has the best performance in line with the waypoints prediction result. The MLP agent can leverage useful features encoded from both RGB and BEV semantic maps. Therefore, the MLP agent can perform as well as the PID agent in estimating steering and throttle. With two different agents considering various aspects of driving, more appropriate action can be decided. Compared to AIM-MT, DeepIPC is better at estimating the steering but worse at estimating the throttle. Yet, it can be said that DeepIPC is better than AIM-MT considering that better steering is more important than better throttle in low-speed driving. Meanwhile, Huang et al.'s model performs the worst as its controller module gets stuck with certain behavior. Be noted that the offline test results can be different from the online test results. This is because any predictions will not affect the next states as they are prerecorded.

\subsection{Online Test}  

The purpose of the online test is to evaluate the model's drivability in driving the vehicle. The model must drive the vehicle safely by following a set of route points while avoiding obstacles (e.g., a vehicle stopped on the left side of the road). The experiment is conducted three times for each condition and on different days to vary the situations. The performance is evaluated based on the average intervention count and intervention time. The less the driver does intervention means the better the driving performance. For a fair comparison, the experiments for all models are monitored by the same driver in preventing a collision. Thus, each intervention is based on the same perspective of the degree of danger. Some driving records can be seen in Fig. \ref{fig:driving_rec}.

Table \ref{tab:drive_result} shows that DeepIPC achieves the best drivability at noon where it has the lowest intervention count and intervention time. Meanwhile, DeepIPC is comparable to Huang et al.'s model in the evening where it achieves the lowest intervention time but has a higher intervention count. Keep in mind that a model with a lower intervention count can have a longer intervention time. For example, a model that fails to make a turn and going to collide needs more correction time than a model that makes a small deviation on a straight path. Hence, it depends on the degree of danger in which the collision is going to happen. Based on the intervention time per intervention count, it is obvious that Huang et al.'s model needs more correction time for each intervention which means that it has the highest danger level compared to DeepIPC and AIM-MT.

Furthermore, in a comparison of drivability in the evening, DeepIPC and AIM-MT perform worse than Huang et al.'s model. In line with the offline result, the model that mainly takes RGB images failed to perceive the environment in the evening as the provided image is not as clearly visible as when driving at noon. On the contrary, Huang et al.'s model become better as it can leverage the information from the depth map that is concatenated with the RGB image from the beginning of the perception phase. This means that although the early fusion strategy causes conflicting features for semantic segmentation, it is useful for driving in low-light conditions. Moreover, even though Huang et al.'s model shows inferior performance on navigational controls estimation in the offline test, its drivability can be said good enough for performing real-world automated driving in the evening with lower traffic compared when driving at noon. Regardless of its comparable performance with DeepIPC in the evening, this exposes the limitation of imitation learning for a model that purely relies on human behavior (by directly predicting steering and throttle levels) without considering another driving aspect that can be obtained from predicting future trajectories in the form of waypoints location in the local coordinate.

\section{Conclusion}
We present DeepIPC, an end-to-end model that can drive a vehicle in real environments. The model is evaluated by predicting driving records and performing automated driving. Furthermore, a comparative study is conducted to justify its performance.

Based on the experimental results, we disclosed several findings as follows. First, in line with our previous work \cite{oskar_tiv}, the BEV semantic feature is proven can improve the model performance in predicting waypoints and navigational controls. With a better perception, the model can leverage useful information which results in better drivability. Second, driving in the low light condition is harder than in the normal condition, especially for DeepIPC and AIM-MT which only rely on RGB images at the early perception stage. Meanwhile, Huang et al.'s model can tackle this issue as it fuses RGB and depth features earlier. Third, considering its performance and the number of parameters in its architecture, DeepIPC can be said as the best model. Lastly, we also validate that the end-to-end imitation learning method is also effective for a complex multi-input multi-output model that is deployed for performing real-world autonomous driving.

As for future works, the perception module can be enhanced with a LiDAR sensor to handle poor illumination conditions such as driving at night. Then, conducting more evaluations on different areas and adversarial situations (e.g., avoiding collision with pedestrians that cross the street suddenly) is suitable to test the drivability further. % 

\bibliographystyle{IEEEtran} 
\bibliography{references.bib}
%\begin{thebibliography}{00}
%\end{thebibliography}

%\vspace{-81pt}

%\begin{IEEEbiography}{Oskar Natan}
\begin{IEEEbiography}[{\includegraphics[width=1in,height=1.25in,clip,keepaspectratio]{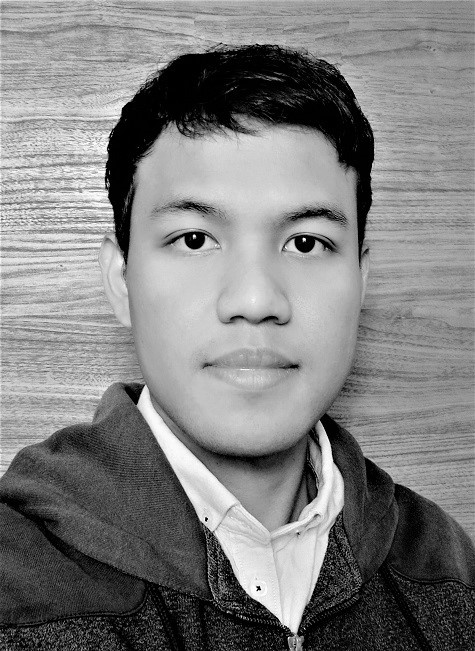}}]{Oskar Natan}
	(Member, IEEE) received his B.A.Sc. degree in Electronics Engineering and M.Eng. degree in Electrical Engineering from Politeknik Elektronika Negeri Surabaya, Indonesia, in 2017 and 2019, respectively. In 2023, he received his Ph.D.(Eng.) degree in Computer Science and Engineering from Toyohashi University of Technology, Japan. Since January 2020, he has been affiliated with the Department of Computer Science and Electronics, Universitas Gadjah Mada, Indonesia, first as a Lecturer and currently serves as an Assistant Professor. His research interests lie in the fields of sensor fusion, hardware acceleration, and end-to-end systems.
	
	Dr. Natan is a member of the Institute of Electrical and Electronics Engineers (IEEE, including the IEEE-ITS Society and IEEE-RA Society) and the Indonesian Computer, Electronics, and Instrumentation Support Society (IndoCEISS). He has been serving as a reviewer for some reputable journals and conferences, including IEEE T-IV, IEEE T-ITS, IEEE ICRA, and IEEE/RSJ IROS. 
\end{IEEEbiography}


%\vspace{101pt}
\vfill

%\newpage

%\begin{IEEEbiography}{Jun Miura}
\begin{IEEEbiography}[{\includegraphics[width=1in,height=1.25in,clip,keepaspectratio]{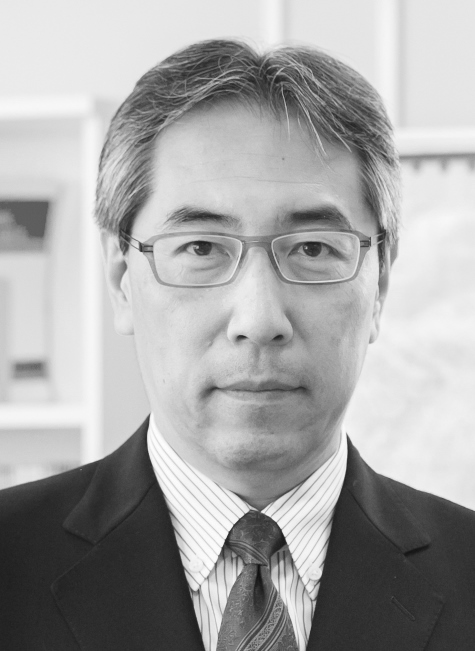}}]{Jun Miura}
	(Member, IEEE) received his B.Eng. degree in Mechanical Engineering and his M.Eng. and Dr.Eng. degrees in Information Engineering from the University of Tokyo, Japan, in 1984, 1986, and 1989, respectively. From 1989 to 2007, he was with the Department of Computer-controlled Mechanical Systems, Osaka University, Japan, first as a Research Associate and later as an Associate Professor. From March 1994 to February 1995, he served as a Visiting Scientist at the Department of Computer Science, Carnegie Mellon University, USA. In 2007, he became a Professor at the Department of Computer Science and Engineering, Toyohashi University of Technology, Japan, where he remains to the present. To date, he has authored or co-authored more than 260 peer-reviewed scientific articles in the field of robotics and autonomous systems in internationally reputable journals and conferences.
	
	Prof. Miura is a member of the Institute of Electrical and Electronics Engineers (IEEE, including the IEEE-RA Society, IEEE-SMC Society, and IEEE-Computer Society), the Robotics Society of Japan (RSJ, Fellow since 2011), the Japanese Society of Mechanical Engineers (JSME), the Japanese Society of Artificial Intelligence (JSAI), the Information Processing Society of Japan (IPSJ), and the Institute of Electronics, Information, and Communication Engineers (IEICE). He has received numerous awards, including the Best Paper Award from the Robotics Society of Japan in 1997, was a finalist for the Best Paper Award at IEEE ICRA in 1995, and the Best Service Robotics Paper Award at IEEE ICRA in 2013. 
\end{IEEEbiography}
%He is also an active member of Institute of Electrical and Electronics Engineers (IEEE) (including IEEE Robotics and Automation Society, IEEE Computer Society, and IEEE Systems, Man, and Cybernetics Society), Robotics Society of Japan (RSJ), Japanese Society for Artificial Intelligence (JSAI), Information Processing Society of Japan (IPSJ), Institute of Electronics, Information, and Communication Engineers (IEICE), and Japanese Society of Mechanical Engineers (JSME).

\vfill

\EOD

\end{document}